\pdfoutput=1
\documentclass{article}

\usepackage{amsmath}
\usepackage{amssymb}
\usepackage{parskip}
\usepackage{amsthm}
\usepackage{graphicx}

\theoremstyle{plain}

\title{De-identification In practice}
\author{Besat Kassaie}
\date{September 2016}
\begin{document}
\maketitle

\begin{abstract}
We report our effort to identify the sensitive information, subset of  data items listed by HIPAA(\textit{Health Insurance Portability and Accountability}), from medical text using the recent advances in natural language processing  and machine learning techniques.  We represent the words with high dimensional continuous vectors learned by a variant of Word2Vec called \textit{Continous Bag Of Words}(CBOW). We feed the word vectors into a simple neural network with a \textit{Long Short-Term Memory} (LSTM) architecture. Without any attempts to extract manually crafted features and considering that our medical  dataset is too small to be fed into neural network,  we obtained promising results. The results thrilled us to think about the larger scale of the project with precise parameter tuning and other possible improvements. 
\footnote{Cheriton School of Computer Science, University of Waterloo}
\end{abstract}
\section{Motivation}
\textit{Electronic Health Record} (EHR) systems have been developed to provide individuals with high quality, continuing health care. These systems are being adopted throughout the world. For example, the Optometry Clinic at the University of Waterloo’s School of Optometry and Vision Science, which is one of the largest vision care centers in \textit{Canada}, adopted an EHR system. In fact, Transparency Market Research has predicted that the value of the global electronic health records solution market to reach 30.28 billion USD by the end of 2023.

As well as supporting individual health care, such data has the potential to be used in a broader scope to improve the lives of multiple generations. Researchers from various domains such as public health, social science, and economics can extract invaluable insights using this data. On the other hand working with medical data requires addressing several problems, such as dealing with unstructured fields, data privacy, and access control. Considering only data privacy, there are a plethora of privacy legislations which have to be considered in any system working on sensitive data, such as \textit{PIPEDA}, \textit{PHIPA}, \textit{FIPPA}, \textit{HIPAA}, and \textit{COPPA}. These concerns make it difficult to share medical data among researchers. 
The Optometry Clinic at UW wishes to give some level of privacy preserving access over their data to optometry researchers . However, due to the lack of ability in ensuring data privacy, currently clinicians can only use data for primary health care, and it is not possible to share the data with researchers. One of the biggest challenge in the  Optometry Clinic at UW data is using free text field in which clinicians are able to write comments. The aim of  these concise and informal comments is to record the patients state, to write some reminders for feature visits or to notify the next clinician about some details of the treatment process. So these free-text fields easily contain many sensitive information such as patient name, the medical centers which the patient is related to, family members identities among others. These types of information not only  make an adversary linkage attack very possible but also in many cases they are not necessary to be included in many of texts. Other issues that have been observed in the optometry data were many misspelled words, extensive use of non standard abbreviations. However, the main objective of the current project is to capture the identifying informations form free text fields ,de-identification, using the state of the art machine learning and Natural Language Processing techniques.

\section{Related Works}

The current project falls into two different area of works ,i.e. automatic free text de-identification and application of deep learning in natural language processing. Extracting specific information from text such as ,in a wide range, from company names to emotions\cite{STR08} is not a new topic. In fact if we consider the de-identification task as a specific kind of Named Entity Recognition (NER) in which the aim is to identify some coarse grain entities such as locations, dates and person it has 25 years of research behind \cite{Rau91},\cite{GR96}. Because of the importance of automatic de-identification of text, a NLP challenge is held as a part of i2b2\textit{Informatic for Integrating Biology \& the Beside} project\cite{i2b2}. The objective of this challenge is the same as ours although the datasets that have been used are different. Some of the approaches that have been applied in this challenge includes classification\cite{Szarvas07}, sequence labeling using CRFs \cite{Wellner07}, rule-based mechanisms\cite{Guillen07} or a hybrid of other approaches \cite{Hara06}. Other works came out which essentially were founded on the challenge approaches such as \cite{Liu15} in which they proposed a hybrid system of rule-based and token/character-level CRFs.  The commonality between all of these works is their intensive works on configuring the problem specific feature sets and capturing context patterns in terms of rules and regular expressions. This makes them very time consuming to develop in real world applications and also hard to generalize them over different datasets. In the machine learning community a large number of works are devoted to extract features and capture the domain automatically. They came up with the idea of using deep neural networks to learn the features and also to retain the context information. There are plenty of NLP tasks  on which variants of deep neural network such as Recurrent Neural Networks  have been applied successfully such as opinion mining \cite{OM14}, speech recognition \cite{hbrid}, language modeling \cite{Zaremba14}, Named Entity Recognition \cite{Hammerton03}, and recently de idenntification of patient notes \cite{Dernoncourt16}. The objective and methods used in our work is quite similar to \cite{Dernoncourt16} from an high level view. Although we use a  different dataset composed of optometry patient records which suffers from  informality. Our way of building the word representation is slightly different as we use a relatively large dataset to learn them, whereas \cite{Dernoncourt16} exploits pre-trained word vectors. Our current LSTM architecture is also different from  \cite{Dernoncourt16}. 

\subsection{Background}

\subsection{Word Representations}
Reflecting semantic and syntactic of words in their representation can improve different natural processing tasks. By relying solely on naive representations such as Bag of Words, i.e. using sparse vectors to represent words as atomic entities, we ignore many levels of semantics and syntactics which are embedded in words and also in their relations with other words.  On the other hand relations such as similarity are not a static and permanent notion and vary based on the context. Recently some interesting works came out which try to capture some levels of semantic and syntactic similarities between words by representing words as continues vectors. We need a representation which can be changed based on the contexts and also can capture the semantics and syntactics existing in an specific context. One approach which has been studied very well so far is using word neighbors. The neighbors can be a large number of words existing in a document or it can be a small window surrounding the word. In one approach based on the window containing the word, we can build a cooccurance matrix in which each entry of the matrix can hold some statistics about the neighbors. The resulting matrix will be sparse and also very large in dimension. For a better performance we need to decrease the dimensionality of the matrix using dimensionality reduction techniques which impose a high computational cost into the process. Other approach is to learn the low dimensional continuous vectors for word representation from the beginning using neural networks. There have been designed various neural network based models to capture this representation \cite{GEH86},\cite{YB03}, \cite{DER86}. Among the proposed techniques we use the model proposed in \cite{TM13}. Two quite similar architectures, are proposed in \cite{TM13} called \textit{Continuous Bag Of Words} and \textit{Continuous Skip-gram}. The former model ,as shown in Figure \ref{CBOW}, uses the continues representation of context, a window surrounding an specific word, to predict the current word. The latter model predicts the context based on the current word. Using the projection learned by the model as the word representation, we are able to capture  semantic and syntactic relations between words in terms of word similarities.
In fact  the learned word representations capture  syntactic and
semantic regularities\cite{TMW13}. Many of these linguistic regularities can be shown as linear  translations between pairs of words. 
The quality of such relations is hard to be evaluated with a rigorous method. It is subjective, intuitive and it firmly depends on the context. However using such representations has shown significant improvements on different downstream NLP tasks.  

\subsection{Conditional Random Fields}

CRFs and semiCRFs\cite{androw} have shown acceptable performance in many areas such as natural language processing, and biomedical domains among others. CRFs are undirected graphical models which provides a discriminative  approach to predict multiple output variables which depend on each other\cite{CHS12}.  Extracting suitable features and introducing them into the model as feature functions is an important an labor intensive step in implementing CRFs.

\subsection{LSTM}
Wherever we face with data which has sequential nature such as text or spoken language the need for capturing the context for each unit of data that can be a word, an audio signal ,or a sentence is crucial. Previous methods
for addressing these problems were using hand-designed
systems  to deal with the sequential nature of data. Recently due to the advances in neural networks, Recurrent Neural Networks (RNN) has been applied extensively on the machine learning learning tasks, related to sequential data. Handwriting recognition \cite{Doetsch14},
\cite{GravesALiwicki}, machine translation \cite{Luong14}, and language
modeling \cite{Zaremba14} are a few among others. In theory standard RNN architectures are able to capture the long term dependencies, however in practice the context for which they are able to maintain dependencies is very small. In fact RNNs suffer from the \textit{vanishing gradient} problem, i.e. sensitivity of the network and subsequently the output to the inputs decrease over time \cite{Alex}.   
LSTMs  \cite{Hochreiter95} are a type of RNNs, that are more effective at capturing long-term temporal dependencies without the optimization problems that comes with standard RNNs. The main idea behind the LSTM architecture is a \textit{memory cell} which can maintain its state over time, and some non-linear
units which regulate the information flow through each LSTM cell. Different architectures of LSTM such as\textit{ Bidirectional Long Short-Term Memory} \cite{A.Graves05} have been designed in which LSTM get trained in both directions over inputs. For the current project we have used an architecture quite similar to the primary LSTM architecture introduced in \cite{Gers00}. To improve the performance of the LSTM, as recommended in \cite{Jozefowicz15}, we add a bias value equal to 1 to the forget gate.

\begin{figure}
	
	\hfill\includegraphics[width=100mm,scale=1]{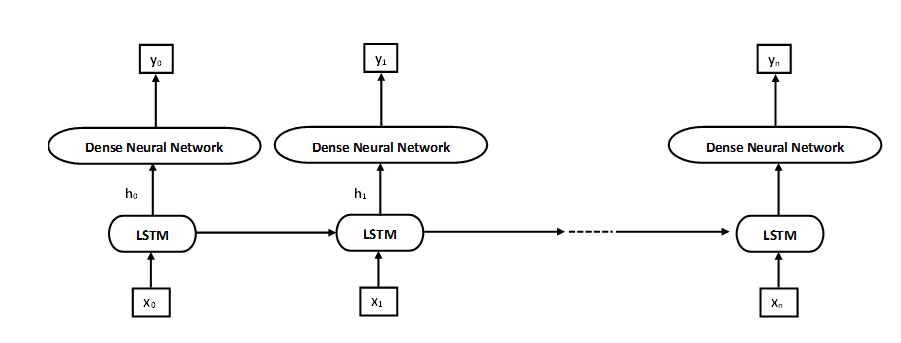}\hspace*{\fill}
	\caption{ Our LSTM architecture which in each time step takes a vector $x_i$ as input and predict the corresponding label $y_i$. In the current model we set $n=15$.  }
	
	\label{LSTM}
\end{figure}


\section{DataSet}
We have used two different datasets for this project: AMIA de-identification challenge dataset \cite{Uzuner07}  and  the Optometry dataset. The reader can look at \cite{Uzuner07} for the description of their dataset.
The optometry dataset contains 68278 records belonging to patients for a period of one year. Each record contains  seven items in  which the practitioner can write comments and notes such as \textit{DIAGNONSIS},\textit{CHIEF-COMPLAINT}, \textit{COMMENT}, \textit{NOTE}, \textit{EXTERNAL-NOTE} among others. Our aim is to identify sensitive information from these seven items. 299 records have been selected randomly to be annotated manually. We view our problem as a sequence labeling problem. We annotated all 299 files in the same way as described in \cite{Uzuner07} to be feed into the CRFs model. We did some extra processes over datasets to train/test our LSTM model as we will describe later. The distribution of tokens are presented in Table \ref{table: disttokopt},\ref{table: distOPTFormal}, and \ref{table: AMIADist}. 
\begin{table}[t! ]
	\begin{tabular}{|p{3.5cm}||p{3.5 cm}|  }
		\hline
	
		Key& Number of Tokens \\
		\hline
		Date   & 1203  \\
		Doctor&   462 \\
		Hospital & 10 \\
		Location    & 258 \\
		Patient&   119  \\
		Phone& 19   \\
		All Identifiers & 2071 \\
		\hline
		
	\end{tabular}
	\caption{Distribution of tokens in Optometry dataset, token refers to the number of words in
		each PHI category.}
	\label{table: disttokopt}
\end{table}

\begin{table}[t! ]
	\begin{tabular}{ |p{3.5cm}||p{3.5 cm}|  }
		\hline
	
		Key & Number of Tokens \\
		\hline
		Date   & 1075  \\
		Doctor&   501 \\
		Hospital & 10\\
		Location    & 2790 \\
		Patient&   985  \\
		Phone& 424   \\
		All Identifiers & 5785 \\
		\hline
		
	\end{tabular}
	\caption{Distribution of tokens in Optometry formal letters dataset.}
	\label{table: distOPTFormal}
\end{table}

\begin{table}[t! ]
	\begin{tabular}{ |p{3.5cm}||p{3.5 cm}|  }
		\hline
		
		Key & Number of Tokens \\
		\hline
		Date   & 7651  \\
		Doctor&   7697 \\
		Hospital & 5204\\
		Location    & 518 \\
		Patient&   1737  \\
		Phone& 271   \\
		All Identifiers & 23078 \\
		\hline
		
	\end{tabular}
	\caption{Distribution of tokens in AMIA dataset.}
	\label{table: AMIADist}
\end{table}
	
\section{Experiments}
HIPAA regulation obliges health care organizations to remove Protected Health Information(PHI) from data sets before sharing them with researchers. HIPAA provides a list of 18 identifiers for the de-identification purpose. Although majority of these categories have been observed in the Optometry dataset, in this project we will focus on the identification  of six categories which have been tried before in the AMIA de-identification challenge held by i2b2 \cite{i2b2}. By focusing on this subset we are able to compare our system performance with the best system \cite{Wellner07} participated in the NLP challenge \cite{Uzuner07} [TODO: need to find i2b2 new dataset 2014-2016]. 

\subsection{Evaluation}
We have used three standard metrics to evaluate our experiments. The metrics are as following:\\\\
$Precision= \frac{True Positive}{True Positive+False Positive}$\\
\\
$Recall= \frac{True Positive}{True Positive+False Negative}$
\\\\
$F-measure= \frac{2* Precision*Recall}{Precision+Recall}$\\

We have computed all of the aforementioned metrics in the token level. For example assume the location PHI  composed of three tokens such as \textit{Canada National Library}. The evaluation can be in instance level or token level. The former means if all of the tokens in PHI are tagged correctly then it is considered as True Positive. Token level considers all of the three words separately.   
Note that as we will explain later we use sliding window approach to generate fixed length segments to feed into LSTM model,  consequently each token falls into more than one segment and  takes different labels. However the metrics are computed on the merged segments to be comparable with the original text. To merge the result (labels) we apply the following rules: 

\begin{itemize}
	\item If all the labels assigned to a token in different windows are the same we keep the label
	\item If the token has different labels, a mixture of non PHI and PHI, we choose PHI label.
	\item If we have any conflicting PHI  we choose the label with larger index. As LSTM uses the past information for prediction,  the  label with larger index is more accurate. 
\end{itemize}

\subsection{Experiment Using CRFs}
As we found that the way of representing the six PHIs in the optometry dataset and the dataset provided by i2b2 is quite similar, we decided  to try the tool \cite{MIST} used by \cite{Wellner07} for de-identification on the optometry dataset. \cite{Wellner07} defines the de-identification as a sequence labeling task for which graphical models can effectively predict the output variables. MIST uses Carafe for sequence labeling. Carafe is a sequence labeling system implemented for phrase identification task and showed a significant performance. In fact Carafe implemented Conditional Random Fields(CRFs) for phrase identification task.
\cite{Wellner07} have introduced some new and task specific features to Carafe to make it applicable on the de-identification task. 

To make sure that the Carafe setting is the same that used in the challenge we first tried the tool on the AMIA data set and produced quite similar results in terms of Precision, Recall and F-measure. A bit of discrepancies exist in our results in compare to the  \cite{Wellner07} due to the some parameter tuning. Then we trained and evaluated their proposed model on the optometry dataset using 3-fold cross validation. By training model, associated weights of feature functions get fitted to optometry data. Surprisingly training MIST CRFs model on the optometry dataset concluded in quit poor results (Table \ref{table: AIMIAOPT}) even in  \textit{Doctor} and \textit{Date} data items which were presented in a very similar way  in both datasets. The results presented in Table \ref{table: AIMIAOPT} are based on token level evaluation.

 The discharge summary  dataset used in the challenge is a set of well written texts. On the other hand \cite{Uzuner07} created this dataset by combining some structured fields so we can see some regularities in each record. In the Optometry free text fields, no regularities can be observed. So we conducted another experiment using referral letter dataset from optometry. We conducted this experiment to investigate whether AMIA's CRFs model, works better for the well written text, as we find in formal letter dataset. The results of this experiment is presented in Table \ref{table: formal letters} showing that the overall result is improved.

 \begin{table}[t! ]
 	\begin{tabular}{ |p{1.5cm}||p{1.5cm}|p{1.5cm}|p{2cm}|  }
 		\hline
 		
 		\hline
 		Key& Precision &Recall&F-measure\\
 		\hline
 		Date   & 0.293    &0.218
 		&  0.25
 		\\
 		Doctor&  0.51
 		 & 0.163
 		    & 0.247
 		    \\
 		Hospital & 0 & 0&  0\\
 		Location    & 0.121
 		 & 0.212
 		 &  0.154
 		 \\
 		Patient&   1
 		  & 1
 		  & 1
 		  \\
 		Phone& 0
 		 & 0
 		   & 0
 		   \\
 		All& 0.321
 		 & 0.321
 		  & 0.275
 		  \\
 		\hline
 
 	\end{tabular}
 	\caption{Running LSTM model on AMIA Test set - Word representations built over Optometry Dataset  }
 	\label{table: OptLSTAMAMIA}
 \end{table}

A main challenge with regard to using CRFs is designing good feature functions. Designing these feature sets is rather straight forward in case of AMIA data. That is firstly because, this dataset is formed from some formally written discharge summaries. Also this unstructured data comes from putting together a number of fields, such as date, hospital name, and etc., from an structured data source. In contrast, the optometry free text data were written in an informal language and many misspelled data as well as abbreviations appear frequently. Also, the source of data in optometry case is unstructured from the beginning. So, due to the lack of regularities in optometry data, manually designing feature functions would be highly challenging. That is why we need other techniques instead of CRFs.

\begin{table}[t! ]
	\begin{tabular}{ |p{1.5cm}||p{1.5cm}|p{1.5cm}|p{2cm}|  }
		\hline
		
		\hline
		Key& Precision &Recall&F-measure\\
		\hline
		Date   & 0.595    & 0.651&   0.621\\
		Doctor&   0.504  & 0.487   &0.495\\
		Hospital & 1.000 & 0.233&  0.333\\
		Location    & 0.741 &0.264&  0.319\\
		Patient&   0.406  & 0.291& 0.319\\
		Phone& 0.963  & 0.054   & 0.092\\
		All& 0.702  & 0.33 & 0.363\\
		\hline
		
	\end{tabular}
	\caption{Training the AMIA model on optometry dataset 3-fold cross validations}
	\label{table: AIMIAOPT}
\end{table}

\begin{table}[t! ]
	\begin{tabular}{ |p{1.5cm}||p{1.5cm}|p{1.5cm}|p{2cm}|  }
		\hline
		
		\hline
		Key& Precision &Recall&F-measure\\
		\hline
		Date   & 0.803    & 0.802&   0.803\\
		Doctor&   0.828 & 0.876   &0.851\\
		Hospital & 0 & 0&  0\\
		Location    & 0.486 &0.462&  0.474\\
		Patient&   0.477  & 0.399& 0.435\\
		Phone& 0 & 0  & 0\\
		All& 0.432 & 0.432 & 0.432\\
		\hline
		
	\end{tabular}
	\caption{Training  LSTM model on optometry dataset 3-fold cross validations}
	\label{table: 9}
	\end{table}

\begin{table}[t! ]
	\begin{tabular}{ |p{1.5cm}||p{1.5cm}|p{1.5cm}|p{2cm}|  }
		\hline
		
		\hline
		Key& Precision &Recall&F-measure\\
		\hline
		Date   & 0.568    & 0.587&   0.576\\
		Doctor&   0.593  & 0.539   &0.564\\
		Hospital & 0 & 0&  0\\
		Location    & 0.467 &0.498&  0.481\\
		Patient&   0.606  & 0.490& 0.541\\
		Phone& 0.575  & 0.583   & 0.579\\
		All& 0.566  & 0.539 & 0.552\\
		\hline
		
	\end{tabular}
	\caption{Training the AMIA model on 160 formal letters - average of, 3 fold cross validation}
	\label{table: formal letters}
\end{table}

\begin{table}[t! ]
	\begin{tabular}{ |p{1.5cm}||p{1.5cm}|p{1.5cm}|p{2cm}|  }
		\hline
		
		\hline
		Key& Precision &Recall&F-measure\\
		\hline
		Date   & 0.833    & 0.966&  0.895 \\
		Doctor& 0.806   & 0.856   & 0.830\\
		Hospital & 0.656  & 0.618 & 0.636 \\
		Location    &0.847  & 0.332& 0.477  \\
		Patient& 1    &1 &1 \\
		Phone & 0.877  & 0.430   & 0.577 \\
		All& 0.8365 & 0.7  & 0.736\\
		\hline
		
	\end{tabular}
	\caption{Training  LSTM model on  AMIA dataset using a composed Word2Vec AMIA+GOOGLE}
	\label{table:  AMIA+GOOGLE }
\end{table}

\subsection{Experiment Using Deep Learning}
\begin{figure}[!ht]
 	\centering
 	
 	\includegraphics[width=50mm,scale=0.5]{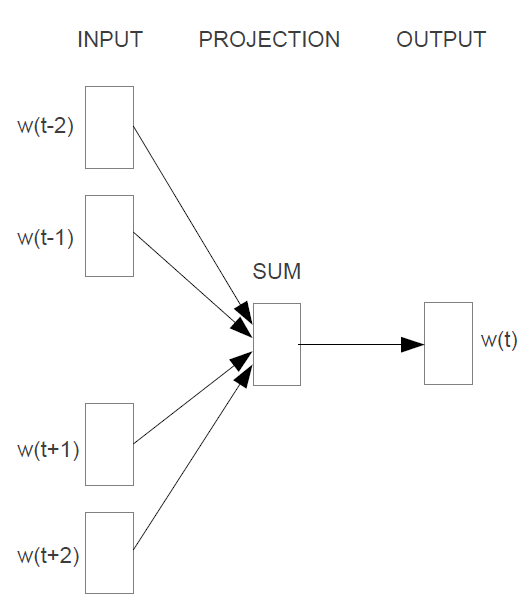}
 	\caption{Continuous Bag of Words \cite{TM13}}
 	\label{CBOW}
 \end{figure}

As RNNs have shown significant performance on sequence labeling tasks in recent studies, we decided to use LSTM, a variant of RNN in our work. LSTM allows us to use automatically extracted features in terms of word embeddings. As distributed continuous word representation is a powerful technique to capture the semantic and syntactics of relations between words, we used the CBOW architecture  as has been shown in Figure \ref{CBOW} to learn word representations. We used all of 68278 records of optometry dataset to build the model. In this architecture we set the window size to five. We ignore words with frequency less than two.  We replace all rare words ,i.e. occurred once, in the training set and unseen words in the test set  with \textit{UNK} \cite{Yao13}. We replace all sequences of numbers  with  \textit{DIGIT} \cite{Collobert11}.  For example \textit{1990}  and \textit{2005} both will be mapped to \textit{DIGITDIGITDIGITDIGIT}. We will try different assignment of \textit{DIGIT} to the numbers. For example we will replace  \textit{1998} with  \textit{DIGIT DIGIT DIGIT DIGIT} to put the burden of learning the pattern on the Word2Vec model shoulder. Also we set the length of word vectors to 200.

\subsubsection{Evaluating Word2vec}
One of the challenges we faced during the project was to evaluate the word2vec model to make sure that  the word vectors are trained well enough to be used in our specific task. We used the test set presented in \cite{TM13}. The test set contains some semantic and syntactic questions. These questions are designed to evaluate different kinds of similarities which exist between words. For example \textit{Italy} and \textit{Canada} are similar in the sense that they are both country. On the other hand \textit{Canada} and \textit{Ottawa} are similar in the sense that \textit{Ottawa} is the capital city of \textit{Canada}. As we can observe there exist different type of similarities between words. In \cite{TM13} they 
designed a set of questions for two pairs of words holding the same similarity relations among. For pairs like (\textit{France}, \textit{Paris}) and (\textit{Canada}, \textit{Ottawa}), a possible question to be asked is  \textit{What is the word that is similar to France in the same way as Canada is similar to Ottawa?}   For answering this question, \cite{TM13} gets the corresponding representation vectors for \textit{Canada}, \textit{Ottawa}, \textit{France}
and by a simple linear transformation such as following  $X= vector(Ottawa)-vector(Canada)+vector(France)$ , they look for a word with highest cosine similarity with vector $X$. In this example $X$ should be $vector(Paris)$.

 We did not know that whether the amount of data used for building word embeddings was sufficient as others, such as \cite{TM13}, have used hundred of millions of words to build their representations. In other words we wanted to realize that whether adding more data will improve the quality of our vectors. 
  
\begin{figure}[!ht]
	\centering

   \includegraphics[width=50mm,scale=1]{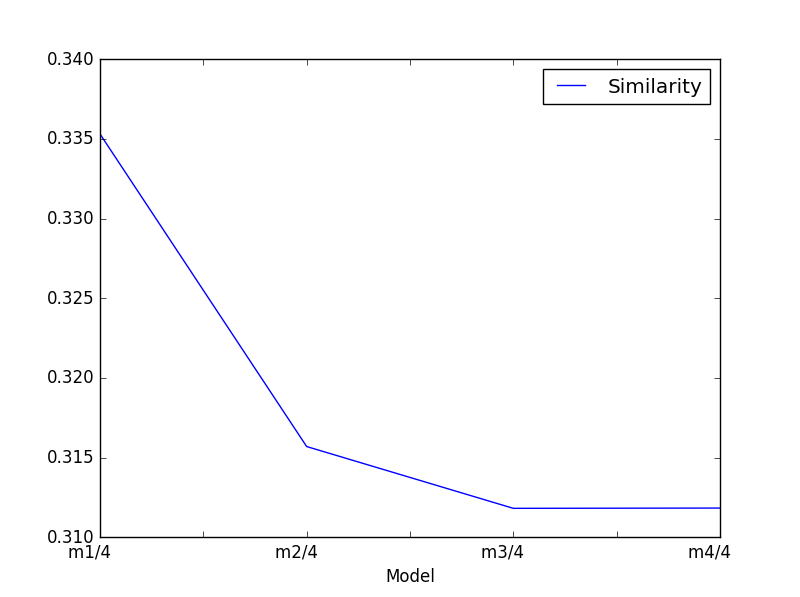}
	\caption{Effect of increasing the data size on the similarity}
	\label{sim}
\end{figure}

 To do so we divided randomly the optometry data set  into four  groups. Then we used the test set(the question list) to evaluate the quality of the Word2Vec model built over 1/4,2/4,3/4, and finally the whole dataset.
To have a measure of quality we fixed the set of questions. We focused on those questions which all of their words exist in 1/4 of our data. For each specific question the amount of cosine similarity between the right and left hand side of the below equation is considered as a quality measure. \\ $vector(word_1)-vector(word_2)+vector(word_4)= vector(word_3) $  \\
\\The cosine similarity is computed for each model. The average of cosine similarities computed for our fixed set of questions over different models are shown in Figure \ref{sim}

 Figure \ref{sim} shows that the average cosine similarity has been decreased as data sized increased from model 1/4 to model 3/4 and then we see that the similarity does not change that much from 3/4 to 4/4. First we wanted to know whether these changes are meaningful. So we need to conduct an statistical test. Our problem statement suit well on an statistical test called\textit{ Repeated Measures }\textit{t} test.  The result of the statistical test has been  shown in Table \ref{table: 1}. Table \ref{table: 1} shows that the increase in data size from model 1/4 to model 2/4 and also from 2/4 to 3/4 has significant impact on the average similarities between pairs. However, we can not say decreasing the similarity could be translated to improvement or worsening of the models.

Our attempt does not show certainly and accurately  that adding more data will improve the quality of the word representations. Since the selected statistical test may not be valid for this case, as by adding more data, we are creating a new model from scratch. In fact we are not capturing the effect of other confounding parameters in our statistical test. Assume that our statistical test is valid, there is no guarantee that the regularities discovered between pairs should be always linear. Finally, we decided to evaluate the quality of word vectors by evaluating the downstream results, after applying LSTM. 
 It is worth to mention that using continues vector representations produces close vectors, in terms of cosine similarity, for misspelled words and abbreviations. For example the most similar word to \textit{pt} was \textit{patient} or the most similar words to \textit{glaucoma} were \textit{glc}, \textit{glauc}, \textit{gluacoma}. This indeed helps to deal with misspelled and abbreviated words.

\begin{table}[t! ]
	\begin{tabular}{ |p{3cm}||p{4cm}|  }
		\hline
		Test& pvalue \\
		\hline
		Model1/4 to 2/4   & 5.3341860080300578e-78   \\
		Model2/4 to 3/4&   1.6130983991735885e-09 \\
		Model3/4 to 4/4 & 0.48765969479743881  \\
		
		\hline
		
	\end{tabular}
	\caption{Repeated Measures t Test results}
	\label{table: 1}
\end{table}

\subsubsection{Data Preprocessing}
For the LSTM Model, we have labeled each word to show that whether it is a part of PHI or not. We have assigned a code for each label. Table \ref{table: 2} shows some of the labels along with the assigned codes.
  
As we need to have a fixed length of word sequences to feed into LSTM, we have created sliding window over the training set. Sliding window may break some of tags. To deal with the broken tags, all the labels in a broken tag replaced with 'O' and only if the tag exists completely inside the window we retain it.

To feed each label into LSTM model it will be replaced by its corresponding code as shown in Table \ref{table: 2}. Also each word get replaced by a  vector with dimensionality of 200, obtained from the Word2Vec model.

\begin{table}[t! ]
	\begin{tabular}{ |p{1.5cm}||p{5cm}|p{4cm}|  }
		\hline

		Label & Code& Decsription\\
		\hline
		BODOC   &[1,0,0,0,0,0,0,0,0,0,0,0,0,0,0,0,0]    & Beginning of Doctor\\
		IODOC&   [0,1,0,0,0,0,0,0,0,0,0,0,0,0,0,0,0] & Inside of Doctor  \\
		BOP&[0,0,1,0,0,0,0,0,0,0,0,0,0,0,0,0,0]& Beginning of Patient\\
		IOP&[0,0,0,1,0,0,0,0,0,0,0,0,0,0,0,0,0] & Inside of Patient \\
		BOD& [0,0,0,0,1,0,0,0,0,0,0,0,0,0,0,0,0]  & Beginning of Date\\
		IOD& [0,0,0,0,0,1,0,0,0,0,0,0,0,0,0,0,0]  & Inside of Date \\
		BOH& [0,0,0,0,0,0,0,0,1,0,0,0,0,0,0,0,0]  & Beginning of Hospital\\
	    O & [0,0,0,0,0,0,0,0,0,0,0,0,0,0,0,0,1]  & Outside\\	
		\hline
		
	\end{tabular}
	\caption{Sample of labels and corresponding codes}
	\label{table: 2}
\end{table}
\subsubsection{Sequence Labeling with LSTM for Optometry Dataset}
We trained LSTM  on the optometry dataset. The architecture of our model is presented in Figure \ref{LSTM}. In this architecture, word sequences of length 15 are fed into the first layer of network, which consists of 15 LSTM component. These word sequences are obtained by applying sliding window over training set as described before.  Each word $x_i$ in these sequences is represented by a vector of length 200. This vector was obtained from the trained Word2Vec model as explained before.

The output of each LSTM component $h_i$ would be another vector of length 200. A dense neural network is attached to each of these output vectors. A dense neural network is simply a fully connected neural network. This network has output vector of length 17 called $y_i$, which represents probability distribution over possible label values. We used dropout of 0.2, \textit{Binary Cross Entropy} for loss function and \textit{Adam}  \cite{Adam14} as our optimization algorithm. We set \textit{epoch} parameter to 10.

In our experiment we used respectively 56\%, 19\% of randomly selected word sequences from optometry data set as training/validation set and  remaining 25\% of sequences in test set. The results is shown in Table {\ref{table: 9}}.

\begin{figure}[!ht]
	\centering
	
	\includegraphics[width=60mm,scale=0.5]{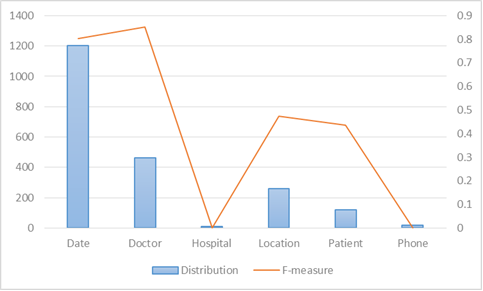}
	\caption{Correlation between the data size and F-measure  }
		\label{DistFmeasure}
\end{figure}

As a second experiment, for evaluating the transferability of the model, learned by LSTM, over optometry dataset, we applied that model, over AMIA dataset. The word representations are the same as what previously we got form Optometry dataset, Table \ref{table: OptLSTAMAMIA} shows the results.

\subsubsection{LSTM for AMIA dataset}

We also tried to use the same strategy of using LSTM to label AMIA dataset. However, in this case we do not have access to a plenty of records for learning word representations based on AMIA dataset. Therefore, we used a set of pretrained word vectors based on news data, provided by Google. Our strategy was to use the few number of records, in AMIA data set, to build a primary vector representation for words. Beside that we also used the vectors, from the pretrained model, as a second source, whenever, we cannot find a word representation based on the primary model, Table \ref{table: NumberWords} shows the distribution of the words obtained from different models. 

Then we applied sliding window approach over the AMIA dataset, similar to what we did with optometry dataset to provide word sequences. We used the same training and test set as  AMIA.  We used the same LSTM network architecture that we used for optometry data. Results are provided in Table \ref{table: OptLSTAMAMIA}. In this experiment we used vector length of 300 as the pre trained model uses this length.

\begin{table}[t! ]
	\begin{tabular}{ |p{3 cm}||p{3 cm}|  }
		\hline

		Word2Vec Model & Number of Words\\
		\hline
	 	GoogleNews    & 1036212  \\
		AMIA Word2Vec &   1281033  \\
		Unknown Words& 23250 \\
	\hline
		
	\end{tabular}
	\caption{Number of word vectors in AMIA test set obtained from different Word2Vec models}
	\label{table: NumberWords}
\end{table}

\section{Discussion}

The abundance of data is a determining factor for applying deep  neural network successfully. However in our case, we had a few number of labeled instances for training our network. This problem is more evident for rare tags in our dataset, such as phone or hospital names, Table \ref{table: disttokopt}. As the results in 
Table \ref{table: 9} shows, for these two tags, our deep learning model was not able to detect any cases. In contrast, the model based on CRFs, was more successful for these tags. The reason is that, in AMIA model, they have introduced very specific features, based on regular expressions and also they have enriched their system by using external lexicons. 

On the other hand, for all the  other tags, for which we had relatively more samples, such as \textit{Date} and \textit{Doctor}, our model outperformed the AMIA model, without using any manually designed feature set. 

The quality of word vectors is another issue in our work. We could not conclude our experiments on quality of our word vectors. Therefore, we are not sure that, given the size of optometry data set, the quality of our word vectors are good enough. This can be assumed as  the another  reason for getting  the results with low accuracy than expected. It is particularly important to note that most such experiments in the literature were conducted over word vectors that were trained over millions of records. A possible remedy for our case could be using the publicly available  pretrained word vectors, such as the one provided by Google, as the initial values instead of other initializations such as random methods , and then refine/adjust  them by training over our own dataset.

Another aspect in our work that has a lot of room for improvement is our network architecture. We have used a preliminary architecture for our neural network, which has just one level of unidirectional LSTM. Many similar works in the literature have used multiple bidirectional layers of LSTM to improve capturing the context in different abstraction levels. Another related and very important issue in our work is setting hyper-parameters. In all steps, from setting the length of sliding windows, to setting network parameters we should try different combination of parameter values. Particularly, setting the hyper-parameters  of the neural network, such as \textit{dropout}, \textit{learning rate} and LSTM specific parameters like \textit{forget rate} are very determining for improving performance. We can employ methods such as grid search for finding the optimum values for the hyper-parameters. Different optimization algorithms as well as different loss functions should be tried. Finally, i2b2 proposed a new dataset for de identification and it worth to conduct our experiments on the new dataset. 

\bibliographystyle{ieeetr}
\bibliography{references}

\end{document}